\documentclass[a4paper,conference]{IEEEtran}
\usepackage{times}
\usepackage{epsfig}
\usepackage{graphicx}
\usepackage{amsmath}
\usepackage{mathrsfs}
\usepackage{amssymb}
\usepackage{booktabs}
\usepackage{multirow}
\usepackage{hyperref}

\DeclareMathOperator*{\argmin}{arg\,min}

\ifCLASSINFOpdf

\else

\fi

\begin{document}

\title{Channel Pruning via Optimal Thresholding}

\author{\IEEEauthorblockN{Yun Ye\IEEEauthorrefmark{1},
Ganmei You\IEEEauthorrefmark{1},
Jong-Kae Fwu\IEEEauthorrefmark{2},
Xia Zhu\IEEEauthorrefmark{2}, 
Qing Yang\IEEEauthorrefmark{1} and 
Yuan Zhu\IEEEauthorrefmark{1}}
\IEEEauthorblockA{\IEEEauthorrefmark{1}Intel,
Beijing, China\\ Email: \{yun.ye, ganmei.you, qing.y.yang, yuan.y.zhu\}@intel.com}
\IEEEauthorblockA{\IEEEauthorrefmark{2}Intel, Santa Clara, USA\\
Email: \{jong-kae.fwu, xia.zhu\}@intel.com}}


\maketitle

\begin{abstract}
Structured pruning, especially channel pruning is widely used for the reduced computational cost and compatibility with off-the-shelf hardware devices. Among existing works, weights are typically removed using a predefined global threshold, or a threshold computed from a predefined metric. The predefined global threshold based designs ignore the variation among different layers and weights distribution, therefore, they may often result in sub-optimal performance caused by over-pruning or under-pruning. In this paper, we present a simple yet effective method, termed Optimal Thresholding (OT), to prune channels with layer dependent thresholds that optimally separate important from negligible channels. By using OT, most negligible or unimportant channels are pruned to achieve high sparsity while minimizing performance degradation. Since most important weights are preserved, the pruned model can be further fine-tuned and quickly converge with very few iterations. Our method demonstrates superior performance, especially when compared to the state-of-the-art designs at high levels of sparsity. On CIFAR-100, a pruned and fine-tuned DenseNet-121 by using OT achieves 75.99\% accuracy with only 1.46$\times$10$^\text{8}$ FLOPs and 0.74M parameters.
\end{abstract}


%
\IEEEpeerreviewmaketitle

\section{Introduction}
\label{section:intro}

Despite the remarkable success in solving many computer vision tasks~\cite{DBLP:journals/pr/GuWKMSSLWWCC18}, deep neural networks (DNNs) are often significantly over-parameterized~\cite{45820}, resulting in redundant and heavy computational work. In practice, the high computation, power and memory requirements always pose challenges when deploying neural networks to low power devices with limited hardware resources, such as mobile phones and Internet of Things/edge devices. In order to facilitates the use of complex neural networks in low power devices, many researches have been carried out to improve the efficiency of DNNs, including efficient architecture design~\cite{DBLP:journals/corr/HowardZCKWWAA17,Zhang_2018_CVPR}, network quantization~\cite{NIPS2015_5647,Hubara:2017:QNN:3122009.3242044}, knowledge distillation~\cite{44873}, low-rank decomposition~\cite{NIPS2014_5544} and network pruning~\cite{NIPS1989_250,DBLP:conf/iclr/0022KDSG17,NIPS2015_5784}.

\begin{figure}[t]
	\begin{center}
		\includegraphics[width=0.9\linewidth]{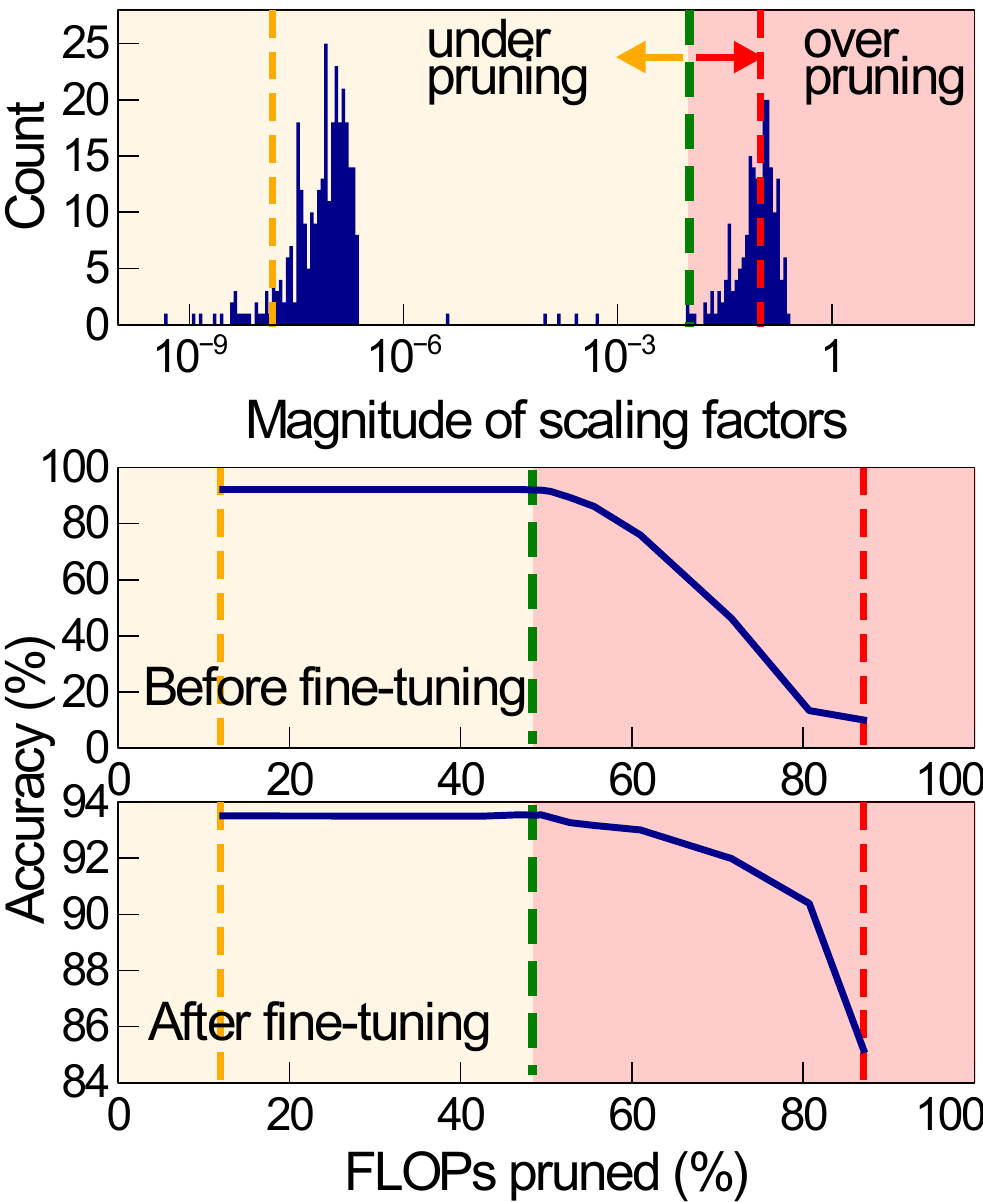}
	\end{center}
	\caption{Illustration of over/under-pruning. \textbf{top}: Histogram of the scaling factors from a batch normalization layer trained with sparsity constraint. The green dashed line depicts the optimal threshold separating the under- and over-pruning region by our method. The orange and red lines are example thresholds of under- and over-pruning respectively. \textbf{middle}: The corresponding accuracy vs. pruned FLOPs of the \textbf{top} plot before fine-tuning, obtained by scaling the optimal thresholds of all layers by certain amounts. \textbf{bottom}: The corresponding plot of the \textbf{middle} plot after fine-tuning.}
	\label{fig:ot_demo}
\end{figure}

Extensive researches have been conducted on network pruning to compress DNNs. Network pruning is popular for its ability to significantly reduce the complexity of a model without or with moderate harm to performance~\cite{8253600}. It can be divided into unstructured pruning and structured pruning. Unstructured pruning~\cite{NIPS2015_5784} directly removes weights without any structural constraints. It can achieve a very high degree of sparsity after pruning. However, specialized hardware devices or libraries~\cite{NIPS2015_5784} are usually required, which limits the pruned models to be deployed on commonly used devices. Structured pruning~\cite{DBLP:conf/iclr/0022KDSG17, NIPS2016_6504,Liu_2017_ICCV} works by removing structured weights, such as filters, layers or branches. It is advantageous in many practical applications because it is friendly to GPU, CPU and many other commonly seen acceleration cards. 

In most existing structured pruning methods, a predefined global threshold is used to determine which part of weights to discard based on an importance metric. For example, Liu \emph{et al.} proposed Network Slimming (NS)~\cite{Liu_2017_ICCV} to impose sparsity constraints on the scaling factors of batch normalization (BN)~\cite{Ioffe:2015:BNA:3045118.3045167} layers. Then, the magnitudes of the scaling factors from all BN layers are sorted. The global threshold is calculated from the sorted magnitudes for pruning a predefined percentage of channels. Such a predefined threshold often results in either under-pruning due to the retention of unimportant weights, or over-pruning due to the discarding of important weights. In the top plot of Figure \ref{fig:ot_demo}, we show the distribution of the scaling factors from the 8$^\text{th}$ BN layer in a VGG-14 model which is trained with $L_1$ penalty following \cite{Liu_2017_ICCV} on CIFAR-10~\cite{Krizhevsky09learningmultiple}. VGG-14 is a modified version of VGG-16~\cite{DBLP:journals/corr/SimonyanZ14a} by replacing the last three fully-connected (FC) layers with a single FC layer and adding BN layers before ReLU~\cite{NIPS2012_4824}. As can be seen from the figure, after training, the scaling factors of a BN layer follow a bimodal distribution, which was also observed in previous works but not carefully discussed~\cite{Liu_2017_ICCV,Gordon_2018_CVPR}. In such a distribution, we assume that the group with larger magnitudes is important, while other weights are negligible. Under this assumption, the optimal threshold should fall in between the important and negligible weights as shown in Figure \ref{fig:ot_demo}. Furthermore, layers in a neural network are generally not equally important to the final performance~\cite{ye2018rethinking}. Therefore, it is natural to expect the pruning threshold is layer dependent.

In this paper, we find that under a wide range of sparsity penalties, the weights are bimodally distributed in most cases. Base on the observations, we assume that there is always an optimal threshold for separating important weights from negligible weights. With this assumption, we propose a very simple, flexible and efficient approach, termed Optimal Thresholding (OT), to prune channels with layer dependent thresholds. To force the channel sparsity, we follow the idea to train the network with $L_1$ penalty on the scaling factors of BN layers, which has been used in recent works~\cite{Liu_2017_ICCV,ye2018rethinking,Gordon_2018_CVPR}. Then, for each layer, we use the cumulative sum of squares as the criterion to find the threshold that optimally separating the scaling factors to prune as many as possible channels with minimal impact on performance. Finally, the pruned model is fine-tuned or trained from scratch (TFS) to form the final compact network. By avoiding over-pruning, OT makes minimal damages to the original model. Therefore the fine-tuning can converge with relatively very few training iterations. Note that though we apply OT layer-wise, it can be naturally extended to more general structures as long as scaling factors are used.

The effectiveness of OT is verified on CIFAR-10/100~\cite{Krizhevsky09learningmultiple} and ILSVRC-2012~\cite{ILSVRC15} with several popular network architectures. OT can reduce the model's size/computations without compromising the performance. It has shown that OT provides even better performance in moderately pruned networks. It can also extend the pruned model to very high sparsity with minimal performance degradation. Compared to the latest state-of-the-art, OT achieves better performance under the same computation budget, especially at the very high sparsity. One-shot pruned models by using OT achieve even better performance than the iteratively pruned models by using NS~\cite{Liu_2017_ICCV}. We also find that TFS achieves better performance than fine-tuning in most cases, which is consistent with~\cite{liu2018rethinking}, except for DenseNet~\cite{Huang_2017_CVPR}. 

The contribution of this paper is three-fold: 1) We propose to prune channels with sparsity determined optimal thresholds for different layers to achieve the best trade-off between complexity and performance; 2) We present Optimal Thresholding (OT), which identify the optimal threshold of the scaling factors by preserving the cumulative sum of squares; 3) We have conducted extensive experiments to demonstrate the superior performance of OT.

\section{Related Works}

As mentioned previously in Section \ref{section:intro}, various works have been done to reduce the computational complexity of neural networks. Efficient architecture design is to reduce the computational complexity with manually designed methods such as depth-wise convolutions~\cite{Chollet_2017_CVPR}, and channel shuffle~\cite{Zhang_2018_CVPR}. Network quantization utilizes lower-precision numerical formats, such as INT8~\cite{Hubara:2017:QNN:3122009.3242044} and binary~\cite{NIPS2015_5647}, to speed up the inference and reduce the memory cost. Knowledge distillation~\cite{44873} transfers knowledge from a large network to a smaller network by using the soft outputs of a large model as targets. Low-rank decomposition~\cite{NIPS2014_5544} approximates the weight matrix as
the product of two low-rank factor matrices with fewer parameters and FLOPs (floating-point operations per second). Network pruning~\cite{NIPS1989_250,NIPS2015_5784} reduces the redundancy by removing insignificant connections from the network. Compared to other approaches, pruning is favored for its ability to reduce the computational cost with relatively minimum performance drop.

\textbf{Structured Pruning.} As mentioned before, due to incompatibility with off-the-shelf hardware for unstructured pruning, structured pruning is preferred in practice. Group Lasso regularization has been utilized for structural sparsity~\cite{NIPS2016_6504}. Li \emph{et al.}~\cite{DBLP:conf/iclr/0022KDSG17} also used the norm of weights ($L_1$-norm) as the metric to prune a predefined percentage of channels but without sparsity constraint. He \emph{et al.}~\cite{He_2017_ICCV} presented a layer-by-layer channel pruning by minimizing the reconstruction error of the next layer. Molchanov \emph{et al.}~\cite{DBLP:conf/iclr/MolchanovTKAK17} applied the Taylor expansion to approximate the channels' effect on loss. A similar idea was explored by propagating the importance score from the final response layer~\cite{Yu_2018_CVPR}. Instead of pruning negligible filters, recent work has also studied pruning redundant or even identical filters~\cite{Ding_2019_CVPR}. 

\textbf{Structured Pruning based on scaling factors}. Research on scaling factor based structured pruning is most closely related to this work. Liu \emph{et al.}~\cite{Liu_2017_ICCV} proposed Network Slimming (NS), which first enforces sparsity on the scaling factors of BN, and then prune channels by a threshold calculated from a predefined drop ratio. The method is very simple to implement but suffers from over/under-pruning which we will demonstrate in Section \ref{sec:sf_distribution}. Ye \emph{et al.}~\cite{ye2018rethinking} studied the same idea. The difference is that Ye adopted ISTA~\cite{doi:10.1002/cpa.20042} to solve the optimization problem with empirically calculated $L_1$ penalties for different layers. To speed up and stabilize the optimization using ISTA, they used a re-scaling trick that scaling filter weights, and BN scaling factors by $1/{\alpha}$ and $\alpha$, respectively ($\alpha<1$). Similar to our method, the channels pruned are layer dependent by using ISTA. But the soft thresholds depend on the learning rate of ISTA in addition to $L_1$ penalties, other than the distribution of scaling factors. Besides, many tricks required for convergence make it much more complicated to implement than NS~\cite{Liu_2017_ICCV}. The idea of sparsifying scaling factors was later extended to more general structures~\cite{10.1007/978-3-030-01270-0_19}, and pruning pre-trained models~\cite{zhonghui2019gate}. Among these methods, an advantage of NS is the simplicity to implement, without using extra modules~\cite{10.1007/978-3-030-01270-0_19,zhonghui2019gate} or special optimization algorithms~\cite{ye2018rethinking,10.1007/978-3-030-01270-0_19}. To the best of our knowledge, none of the aforementioned work has carefully studied how to choose the pruning threshold considering the inter- and intra-channel distribution of scaling factors.

\section{Methodology}

\begin{figure}[t]
	\begin{center}
		\includegraphics[width=0.95\linewidth]{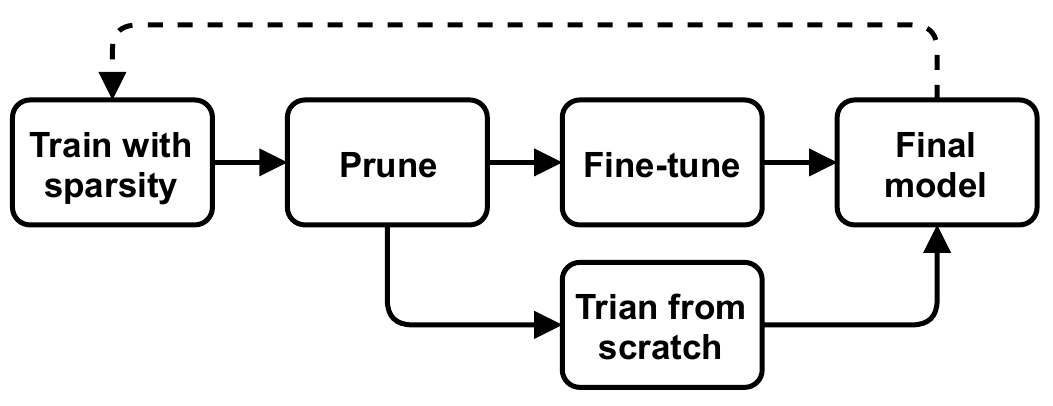}
	\end{center}
	\caption{A typical pipeline of pruning with sparsity constraint.}
	\label{fig:prune_pipeline}
\end{figure}

\subsection{Training with Sparsity}
A typical pipeline we follow is illustrated in Figure \ref{fig:prune_pipeline}. In the pipeline, we first solve the following sparsity constrained problem:

\begin{equation} \label{eq:loss0}
\hat {\theta}=\argmin_{\theta} \mathcal{L}\left(\theta\right)+\mathcal{R}\left(\theta\right),
\end{equation}

\noindent
where $\mathcal{L}$, $\mathcal{R}$, and $\theta$ are loss function, regularization term for sparsity, and trainable parameters, respectively. The regularization term aims to reduce the complexity under the constraints of model size, computations and storage/memory cost. In this work, we follow the idea of pruning channels since the number of channels is strongly correlated to the model complexity and it is very easy to implement in practical use cases. Specifically, we impose sparsity on the scaling factors of BN layers by using $L_1$ regularization, which is a common method to force sparsity~\cite{NIPS2015_5784,Liu_2017_ICCV}. Therefore Equation \ref{eq:loss0} turns out to be

\begin{equation} \label{eq:loss1}
\hat {\theta}=\argmin_{\theta} \mathcal{L}\left(\theta\right)+\lambda\sum_{\gamma \in \Gamma}\left|\gamma\right|,
\end{equation}

\noindent
where $\Gamma$ is a set of scaling factors in the target neural network, while $\lambda$ controls the degree of sparsity. $\Gamma$ can either be the set of all the scaling factors or the set of scaling factors of the specified BN layers.

\subsection{Distribution of Scaling Factors}
\label{sec:sf_distribution}
Gradient descent with $L_1$ regularization acts like soft-thresholding that the $\gamma$s without strong gradients against it will be close to zero (around or smaller than $lr\times\lambda$) after training. Other $\gamma$s with strong gradients will remain much larger values. Thus after training the $\gamma$s tend to be bimodally distributed. As pointed out by ~\cite{ye2018rethinking}, the importance of channels can be measured by $\gamma$s. We assume that the group of larger $\gamma$s is important, while others are negligible. 

In addition to the under/over-pruning problem, we demonstrated in Section \ref{section:intro}, the difference in distribution between layers is also a problem for the predefined threshold. As an example, we plot the histogram of scaling factors of the 7$^\text{th}$ and the 10$^\text{th}$ BN layers from the same VGG-14 model. As shown in the figure, the groups of important weights are not aligned. By using a global threshold, it will be problematic for layers with smaller scaling factors. For example, if we want to prune 73\% of the channels following NS~\cite{Liu_2017_ICCV}, the calculated threshold is good for the 7$^\text{th}$ BN layer but is too big for the 10$^\text{th}$ layer resulting in over-pruning the layer. This also limits the higher pruning percentage because the removal of the whole 10$^\text{th}$ BN layer will break the network, which is made of sequentially stacked convolutional layers. 

\begin{figure}[t]
	\begin{center}
		\includegraphics[width=0.9\linewidth]{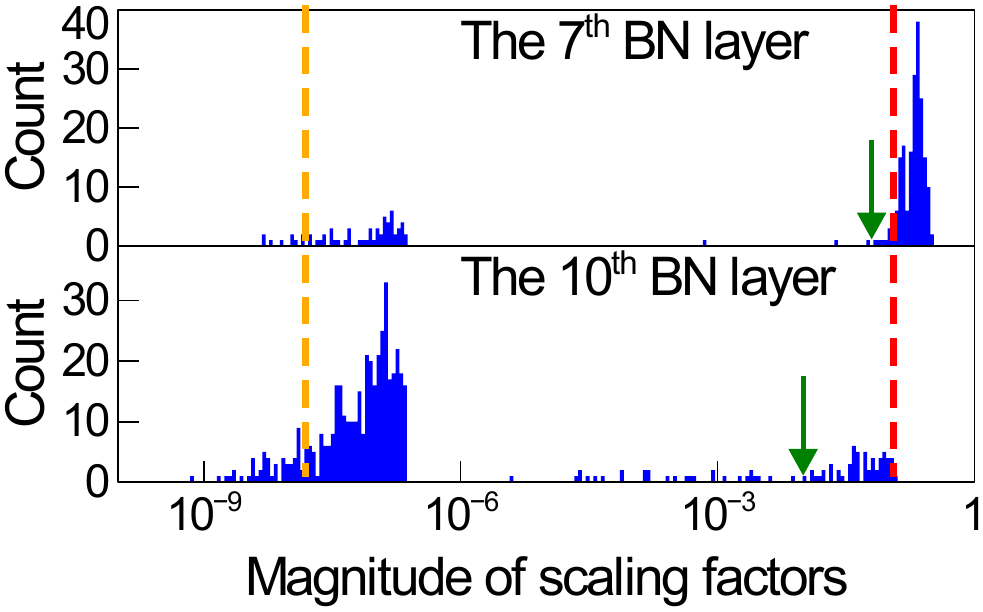}
	\end{center}
	\caption{Distributions of scaling factors in different BN layers. The left (orange) and right (red) dashed lines represent the thresholds for pruning 30\% and 73\% percent of channels respectively. The green arrows are the thresholds identified by OT.}
	\label{fig:sf_layer_dist}
\end{figure}

\begin{figure}[t]
	\begin{center}
		\includegraphics[width=0.9\linewidth]{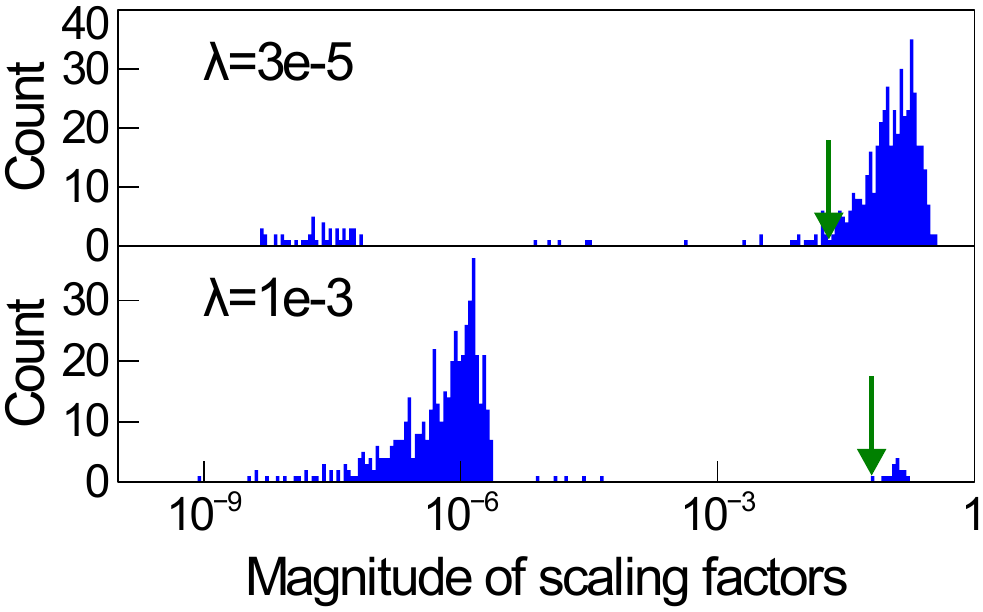}
	\end{center}
	\caption{Distributions of scaling factors under different $L_1$ penalties. The green arrows are the thresholds identified by the proposed method. The 8$^\text{th}$ BN layer of VGG-14 is used for the plots.}
	\label{fig:sf_sparsity_dist}
\end{figure}

To determine the optimal threshold for a specific $L_1$ penalty, we also studied the distribution of scaling factors under different intensities of $L_1$. We observed that in the range from very weak to strong $L_1$ regularization, the important scaling factors are always several orders of magnitude larger than the negligible factors. As an example in Figure \ref{fig:sf_sparsity_dist}, we show the distribution of scaling factors from the 8$^\text{th}$ BN layer of VGG-14, under $\lambda$=3e-5 and $\lambda$=1e-3. 

\subsection{Optimal Thresholding for Pruning}

To handle different BN layers and different sparsity levels, a principled method is needed to achieve the best trade-off between efficiency and performance. For a given set of scaling factors $\Gamma$, we denote $\Gamma_N$ as the subset of negligible $\gamma$s, and $\Gamma_I$ as the subset of important $\gamma$s. Here, we consider the most common case that both $\left|\Gamma_I\right|$ and $\left|\Gamma_N\right|$ are not empty. Ideally the optimal threshold is a $\gamma_{th}$ such that ${\sup \Gamma_N}<\gamma_{th}\le{\inf \Gamma_I}$. For $p\ge0$, the following inequalities hold:

\begin{equation} \label{ineq:sum_np}
\sum_{\gamma \in \Gamma_N}\left|\gamma\right|^p\le\left|\Gamma_N\right|\left|\sup \Gamma_N\right|^p\
\end{equation}
\begin{equation} \label{ineq:sum_ip}
\left|\Gamma_I\right|\left|\inf \Gamma_I\right|^p\le\sum_{\gamma \in \Gamma_I}\left|\gamma\right|^p\le\left|\Gamma_I\right|\left|\sup \Gamma_I\right|^p.
\end{equation}

\noindent
Then we have
\begin{equation}
\begin{aligned}
\frac 
{\sum_{\gamma \in \Gamma_N}\left|\gamma\right|^p}
{\sum_{\gamma \in \Gamma}\left|\gamma\right|^p}
\le
\frac 
{\sum_{\gamma \in \Gamma_N}\left|\gamma\right|^p}
{\sum_{\gamma \in \Gamma_I}\left|\gamma\right|^p}
\le
\\\frac 
{\left|\Gamma_N\right|\left|\sup \Gamma_N\right|^p}
{\left|\Gamma_I\right|\left|\inf \Gamma_I\right|^p}
=\frac {\left|\Gamma_N\right|}{\left|\Gamma_I\right|}\alpha^{-p},
\end{aligned}
\end{equation}

\noindent
and

\begin{equation}
\begin{aligned}
\frac 
{\sum_{\gamma \in \Gamma_N}\left|\gamma\right|^p
	+\left|\inf\Gamma_I\right|^p}
{\sum_{\gamma \in \Gamma}\left|\gamma\right|^p}
\ge
\frac
{\left|\inf\Gamma_I\right|^p}
{\sum_{\gamma \in \Gamma}\left|\gamma\right|^p}
\ge
\\\frac
{\left|\inf\Gamma_I\right|^p}
{\left|\Gamma\right|\left|\sup \Gamma_I\right|^p}
=
\frac
{1}
{\left|\Gamma\right|}\beta^{-p},
\end{aligned}
\end{equation}

\noindent
where $\alpha=\frac{\inf \Gamma_I}{\sup \Gamma_N}$, and $\beta=\frac{\sup \Gamma_I}{\inf \Gamma_I}$. So if we can find a $\delta$ such that $\frac{\left|\Gamma_N\right|}{\left|\Gamma_I\right|}\alpha^{-p}\le\delta\le\frac{1}{\left|\Gamma\right|}\beta^{-p}$, it can be used to separate important and negligible $\gamma$s. In our experiments, by ignoring the very small portion of outliers falling in between the two subsets, we observed that $\alpha$ ranges from $10^5$ to $10^6$ and $\beta$ ranges from 2 to 100 typically, and the higher $\lambda$ is the lower $\beta$ is. The term $\frac{\left|\Gamma_N\right|}{\left|\Gamma_I\right|}$ and $\frac{1}{\left|\Gamma\right|}$ are bounded by the design of the network, and they do not change with $p$. Therefore, as long as $p$ is large enough, we will have $\frac{\left|\Gamma_N\right|}{\left|\Gamma_I\right|}\alpha^{-p}<<\frac{1}{\left|\Gamma\right|}\beta^{-p}$, even for comparing $\Gamma$ from different convolution layers. Based on this, we can simply use $\frac{1}{\left|\Gamma\right|}\beta^{-p}$ as an estimation of $\delta$ to calculate the threshold.

In practice we find that $p$=2 is enough to find the threshold. Specifically, define $\Gamma_<\left(x\right)=\left\{\gamma|\gamma \in \Gamma, \gamma < x\right\}$. Then we find the optimal threshold as

\begin{equation} \label{eq:ot}
\begin{split}
\gamma_{\mathrm{th}} & =\gamma' \in \Gamma, \\
\mathrm{s.t.} & \sum_{\gamma \in \Gamma_<\left(\gamma'\right)} \gamma^2 
<
\delta \sum_{\gamma \in \Gamma}{}\gamma^2 \le
\sum_{\gamma \in \Gamma_<\left(\gamma'\right)} \gamma^2+{\gamma'}^2.
\end{split}
\end{equation}

\noindent
$\gamma_{\mathrm{th}}$ can be easily calculated by 1) sorting the scaling factors in ascending order and then 2) finding the first scaling factor with a cumulative sum of squares larger than or equal to $\delta$.

In practice, we observed that this algorithm can reliably find the threshold, as we would expect in various situations. For examples, the optimal thresholds found by Equation \ref{eq:ot} are shown as the green dashed line in the upper plot in Figure \ref{fig:ot_demo}, and the arrows in Figure \ref{fig:sf_layer_dist} and Figure \ref{fig:sf_sparsity_dist}. 

With $\gamma_{\mathrm{th}}$ we then proceed to the pruning step in the pipeline shown in Figure \ref{fig:prune_pipeline} by dropping the channels where the scaling factors are smaller than $\gamma_{\mathrm{th}}$ layer-wise. In a sequentially stacked convolutional network, a BN layer is usually located between of two convolutional layers, so the output channels of the preceding layer and the input channels of the subsequent layer should be pruned accordingly. The advantage over NS is that OT does not suffer from breaking a network by pruning the entire layer. After pruning, the model can be further fine-tuned or re-trained from scratch if enough computation resources are available. The process can be iteratively repeated as illustrated in Figure \ref{fig:prune_pipeline}.

\textbf{Dealing with skip connections}. In addition to the common architectures with sequentially stacked convolutional layers, OT can also be applied to modern architectures with skip connections, such as ResNet~\cite{He_2016_CVPR} and DenseNet~\cite{Huang_2017_CVPR}. Unlike sequentially stacked convolutional layers, a network with skip connections behaves like ensembles of many relatively shallow networks~\cite{NIPS2016_6556}. This property brings much better robustness to the lesion as discussed in \cite{NIPS2016_6556}. The main difference compared to a sequentially stacked convolutional network is that the removal of an entire BN layer may not break the network. To handle the possible redundant branch, we first calculate a global optimal threshold $\gamma_{\mathrm{g}}$ by applying Equation \ref{eq:ot} to all the scaling factors of the network. Then, for all the branches with BN layers, we remove the entire branch if the scaling factors of the last BN layer are smaller than $\gamma_{\mathrm{g}}$. 

In some networks with skip connections, the BN layers sharing input channels with other parallel branches cannot be pruned trivially. For this type of BN layer, we follow the approach in NS~\cite{Liu_2017_ICCV} by using channel selection operation to mask out the negligible channels.

\section{Experiments}

Our method was evaluated on three datasets including CIFAR-10~\cite{Krizhevsky09learningmultiple}, CIFAR-100 and ILSVRC-2012~\cite{ILSVRC15} with popular network architectures. We compared the results with NS~\cite{Liu_2017_ICCV} and Ye \emph{et al.}~\cite{ye2018rethinking}. Both our method and the reproduction of NS are implemented using PyTorch~\cite{paszke2017automatic}. For simplicity, we use $\delta$=1e-3, which is estimated from the typical values based on our experimental settings, for all the experiments. 

\subsection{Experimental Setup for CIFAR-10/100}
CIFAR-10 \& CIFAR-100 are two classification datasets with 10 \& 100 classes, respectively. Each dataset includes 60000 32$\times$32 color images. Both datasets are split to 50,000 train images and 10,000 test images. In our experiments, we pad each image with 4 pixels, then randomly crop and mirror the image for data augmentation.

To train with sparsity, we follow the same setup of NS. Stochastic Gradient Descent (SGD) with a Nesterov momentum of 0.9 and weight decay of 1e-4 is used for optimization. The network is trained for 160 epochs. The initial learning rate is 0.1, and is decayed by a factor of 10 at the 80$^\text{th}$ and 120$^\text{th}$ epoch. The batch size is 64. The scaling factors of all BN layers are initialized to 0.5 as suggested by \cite{Liu_2017_ICCV}.

For CIFAR, we modified three popular architectures for evaluation. In addition to the VGG-14 mentioned earlier, we obtained ResNet-50 and DenseNet-121 for CIFAR both by 1) Change the kernel size, stride, and padding of the first convolution to 3$\times$3, 1 and 1 respectively; 2) Remove the max-pooling layer. We also re-implemented ResNet-20 for CIFAR-10 following \cite{ye2018rethinking}. The $\lambda$s used for these architectures are summarized in Table \ref{table:lambdas}.

\begin{table}[b]
	\begin{center}
		\setlength{\tabcolsep}{2.5pt}
		\begin{tabular}{c|c|c} 
			\toprule[2pt]
			Architecture & Dataset & $\lambda$s \\
			\midrule[1pt] 
			\multirow{2}*{VGG-14} 
			& CIFAR-10 & \{1,2,5\}e-5,\{1,2,5\}e-4,1e-3 \\ 
			& CIFAR-100 & 5e-5,\{1,2,5\}e-4,1e-3 \\  
			\midrule[0.5pt]
			\multirow{2}*{ResNet-50} 
			& CIFAR-10 & \{2,5\}e-5,\{1,2,3,4,5\}e-4 \\ 
			& CIFAR-100 & 5e-5,\{1,2,3,4,5\}e-4 \\
			\midrule[0.5pt]
			\multirow{2}*{DenseNet-121} 
			& CIFAR-10 & \multirow{2}*{\{1,2,5\}e-5,\{1,2\}e-4} \\
			& CIFAR-100 &  \\
			\midrule[0.5pt]
			ResNet-20 & CIFAR-10 & 8e-4,\{1,2,3,4\}e-3 \\
			\bottomrule[2pt] 
		\end{tabular}
	\end{center}
	\caption{$\lambda$s for training different architectures on CIFAR dataset.}
	\label{table:lambdas}
\end{table}

After pruning, the models are further fine-tuned for 40 epochs with a learning rate of 0.001. We found this to be sufficient for both NS and OT to converge. Especially for OT, we observed that it converges in less than 3 epochs for most cases. We also conducted experiments by TFS for the pruned models and we follow the same amount plan for computation budget as suggested by \cite{liu2018rethinking}. If the pruned model saves less than 2$\times$ FLOPs, we train with epochs that achieve the same FLOPs used in the original training, otherwise, we double the number of epochs.

\subsection{Experimental Setup for ILSVRC-2012}
ILSVRC-2012 is a large scale image dataset of 1,000 classes. We experimented with 1,281,167 training images and 50,000 validation images. We train the model for 90 epochs and fine-tune pruned models for 15 epochs with a batch size of 256. The same SGD settings and initial learning rate with CIFAR are used. The learning rate is decayed by a factor of 10 at the 30$^\text{th}$ and 60$^\text{th}$ epoch. The training data is augmented by random resized cropping to 224$\times$224 and horizontal mirroring. For random resized cropping, the size ratio is randomly sampled between 0.08 and 1.0, and the aspect ratio is sampled from 3/4 to 4/3. Validation images are resized to 256$\times$256, then center cropped to 224$\times$224. We report the accuracy on the validation set. We also report the accuracy after 1 epoch fine-tuning to check if the model converges quickly. VGG-16-BN and PreResNet-50 \cite{10.1007/978-3-319-46493-0_38} are used for experiments. VGG-16-BN is modified from VGG-16 by adding BN layers before ReLU, and replacing Dropout~\cite{JMLR:v15:srivastava14a} layers by BN layers. For OT, we set $\lambda$=2e-4 for both to achieve a relatively high sparsity. For NS, $\lambda$=1e-5 for VGG-16 and $\lambda$=1e-4 for PreResNet-50 are used following \cite{Liu_2017_ICCV}, and the percentages of pruned channels are calculated to achieve the same (or a little bit higher for convergence) computation budgets with models pruned by OT.

\subsection{Verifying the Optimal Thresholds}

We first verified the optimality of the thresholds determined by our method. For a VGG-14, after trained on CIFAR-10 with $\lambda$=1e-4, we calculate the thresholds of all layers by OT. Then, we shift the thresholds of all layers simultaneously by a number in log10 scale. The shift is sampled from -6 to 0.5 to cover a reasonable and sufficient range for both under- and over-pruning. The results are shown in the middle and lower graphs of Figure \ref{fig:ot_demo}. The middle graph is the test accuracy before fine-tuning, we can see that in the under-pruning region (pale orange), the pruned percentage of FLOPs increases without any degradation in accuracy. This indicates that the pruned channels are all negligible for the final performance. When the threshold shifts into the over-pruning region (pale red), a distinct performance drop is observed, implying that importance weights are removed. A similar trend can also be observed in the lower graph, which is the result of fine-tuning. Although fine-tuning can partially recover the accuracy, the performance is lower than the optimal threshold in the over-pruning region.

\begin{figure*}[t]
	\begin{center}
		\includegraphics[width=0.9\linewidth]{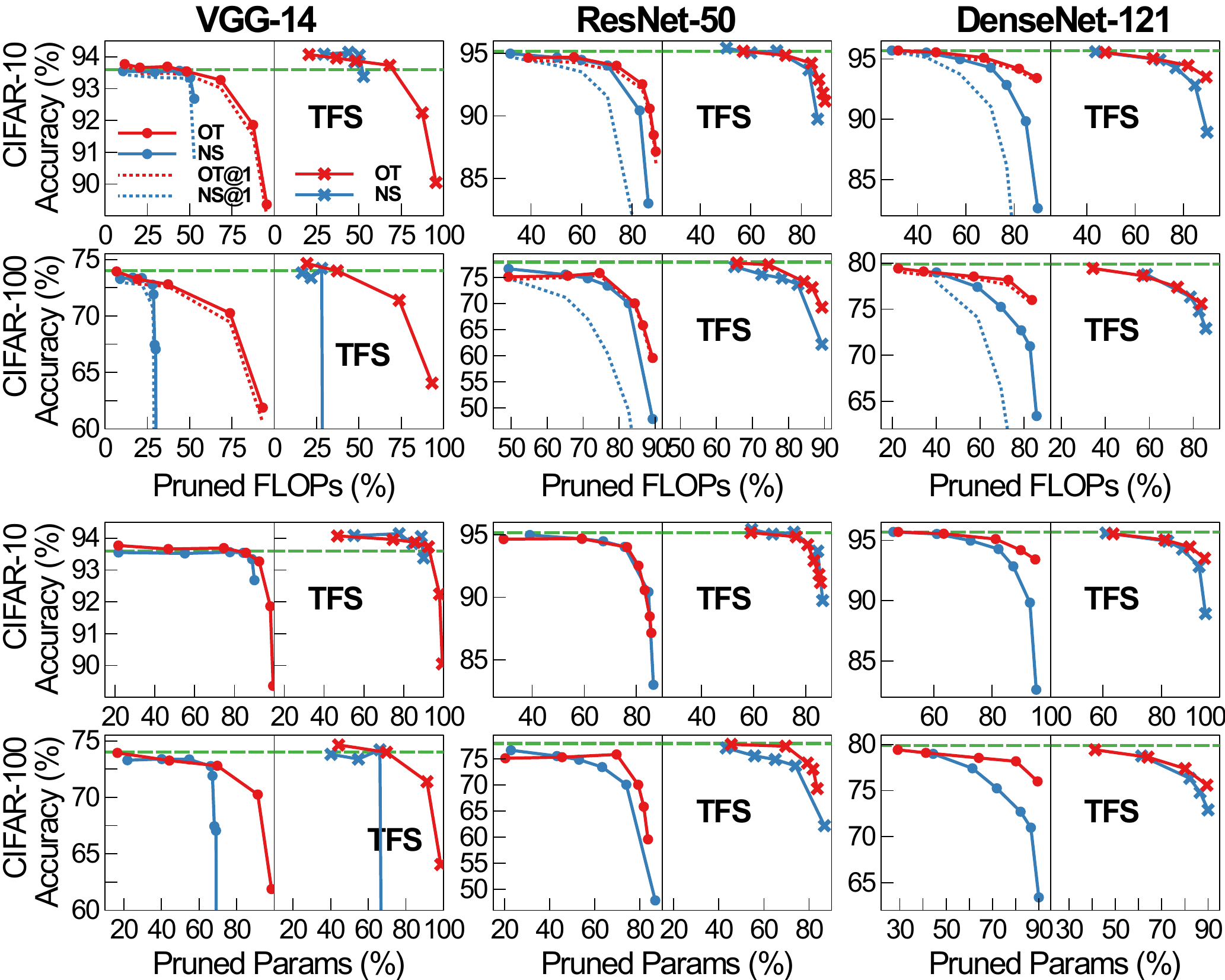}
	\end{center}
	\caption{Results on CIFAR-10/100, compared with NS. The three columns are results for VGG-14, ResNet-50 and DenseNet-121 respectively. The first two rows are accuracy vs. percentage of pruned FLOPs, and the last two rows are accuracy vs. percentage of pruned parameters. In each column, the left plots are results by fine-tuning and the right plots are results by TFS. The green dashed line represents the baseline accuracy. The dotted line represents the accuracy fine-tuned by 1 epoch. All plots share the same legend as in the left top plot.}
	\label{fig:cifar_fttfs}
\end{figure*}

\begin{figure*}
	\begin{center}
		\includegraphics[width=0.9\linewidth]{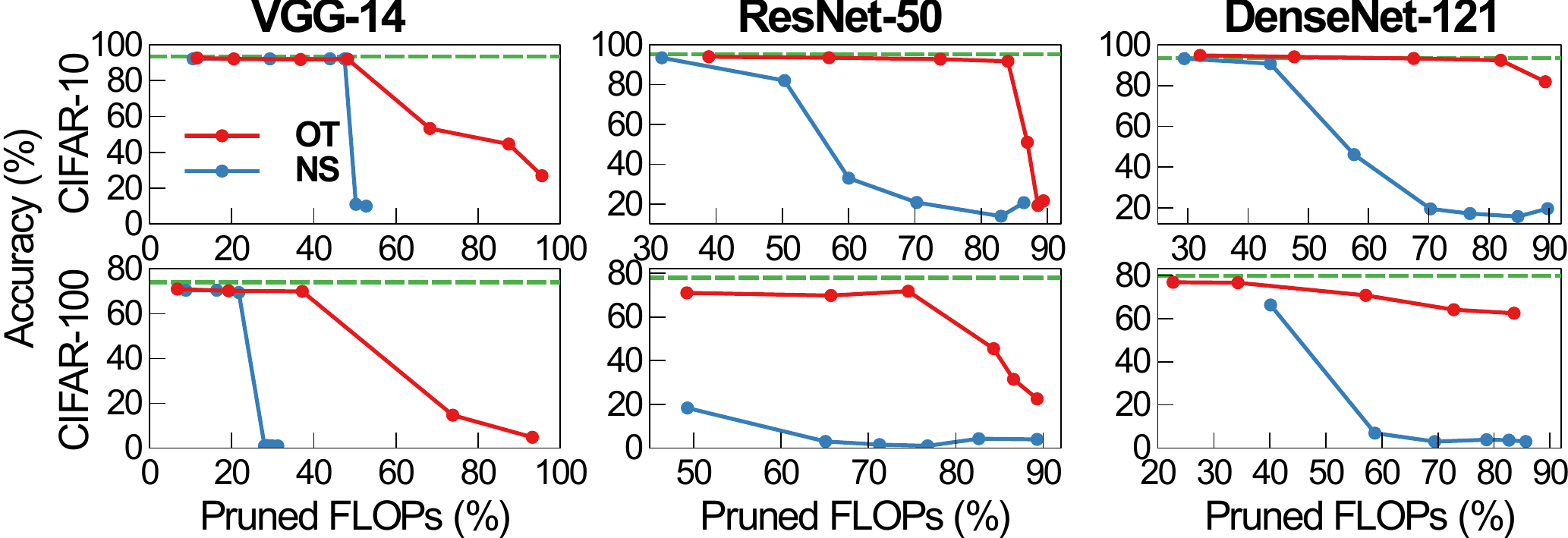}
	\end{center}
	\caption{Accuracy vs. pruned FLOPs without any further training. The three columns are for VGG-14, ResNet-50 and DenseNet-121 respectively. All plots share the same legend as in the left top plot. The green dashed line represents the accuracy of the baseline model.}
	\label{fig:cifar_prune}
\end{figure*}

\subsection{Results on CIFAR-10/100}

In this section, we evaluated OT on CIFAR-10/100 and compared OT with NS. For NS, we use $\lambda$=1e-4 for VGG-14, and $\lambda$=1e-5 for ResNet-50 and DenseNet-121 as  recommended in the original paper. Then, different prune ratios are applied to get pruned networks. For OT, we train with different $\lambda$s for the pruned networks. Note that although we trained with many different $\lambda$s for OT, this does not imply that OT requires more computations on training with sparsity than NS. For NS, the optimal $L_1$ penalty is obtained by performing a grid search on training with different $\lambda$s as described in the original paper. 

\begin{table}[b]
	\label{table:model_summary}
	\begin{center}
		\begin{tabular}{c|c|c|c} 
			\toprule[2pt]
			Architecture & FLOPs & \ Params (M) & Accuracy \\
			\midrule[1pt] 
			\multirow{2}*{VGG-14} 
			& 314.59M & 14.73 & 93.59\% \\ 
			& 314.68M & 14.77 & 74.0\% \\  
			\midrule[0.5pt]
			\multirow{2}*{ResNet-50} 
			& 1.25G & 23.52 & 95.15\% \\ 
			& 1.25G & 23.71 & 77.94\% \\
			\midrule[0.5pt]
			\multirow{2}*{DenseNet-121} 
			& 908.21M & 6.96 & 95.67\% \\
			& 908.39M & 7.05 & 79.87\% \\
			\bottomrule[2pt] 
		\end{tabular}
	\end{center}
	\caption{Details of the baseline models. In each cell the first row is for CIFAR-10 and the second row is for CIFAR-100. M and G represent $10^6$ and $10^9$ respectively}
\end{table}

The results are shown in Figure \ref{fig:cifar_fttfs}. The corresponding reference FLOPs, the number of parameters and the baseline accuracies are listed in Table \ref{table:model_summary}. The baseline accuracies listed in Table \ref{table:model_summary} are drawn in green dashed line in each figure. In Figure \ref{fig:cifar_fttfs}, we can see that OT consistently outperform NS in pruning a high percentage of FLOPs for both fine-tuning and TFS, which indicates the advantage of avoiding over-pruning. Especially for VGG-14, only up to 53\% FLOPs for CIFAR-10 and 31\% FLOPs for CIFAR-100 can be pruned because more pruning will break the network by removing an entire layer. In most cases, OT is better than NS in pruning parameters, except that ResNet-50 trained on CIFAR-10 can be comparable to NS. 

By using OT, we have similar observations with \cite{liu2018rethinking} that TFS has better performance than fine-tuning in most cases, except for DenseNet. A distinct difference between fine-tuned and TFS results for DenseNet-121 is not seen. We conclude that this phenomenon may be related to both over-pruning and the ubiquity of skip-connections, which requires further study.

\begin{figure}[t]
	\begin{center}
		\includegraphics[width=0.9\linewidth]{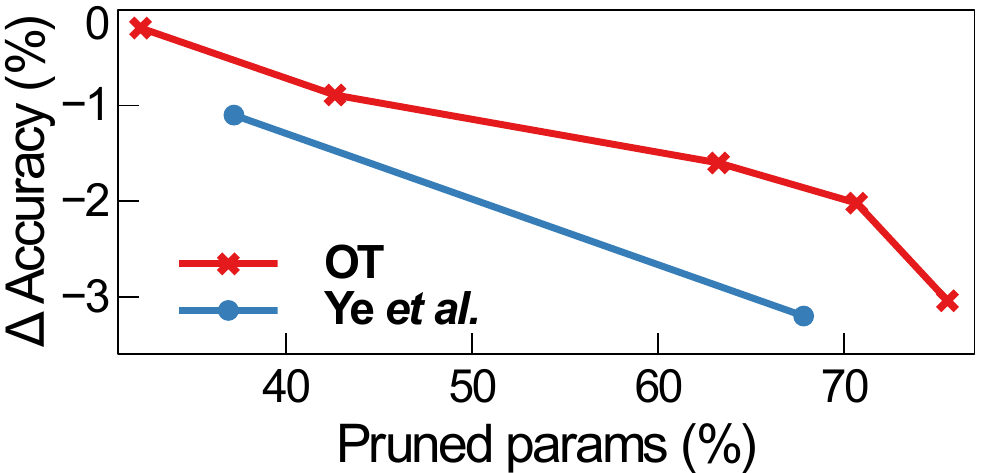}
	\end{center}
	\caption{Accuracy drop of ResNet-20 on CIFAR-10.}
	\label{fig:iclr2018}
\end{figure}

In addition to the results of fine-tuned and TFS models, we also checked the impact on the networks by showing the accuracy of the pruned models without any training in Figure \ref{fig:cifar_prune}. It shows that NS significantly destroys the original networks through over-pruning, while OT induces relatively much less damage to the original networks. In addition, we observed that for both OT and NS, DenseNet is the least sensitive architecture for pruning with the most skip-connections. The result suggests that skip-connection improves the robustness of convolutional neural networks. 

As mentioned earlier, one of the advantages of using OT is that the pruned model can quickly converge during fine-tuning with very few iterations. We show this in the first two rows of Figure \ref{fig:cifar_fttfs}. In the figure, \textbf{OT@1} and \textbf{NS@1} are the accuracy fine-tuned by 1 epoch using OT and NS respectively. By only 1 epoch training, OT is very close to the result obtained by training 40 epochs, while the NS accuracy after 1 epoch is much worse than the final result, making fine-tuning more time consuming.  

In Table \ref{table:statistics} we compare the architectures by channel statistics over 11 runs, on CIFAR-10 and CIFAR-100 respectively. For each run we first train VGG-14 with $\lambda$=2e-4, then the model is pruned and fine-tuned. Note that we intend to keep the same number of channels for comparison, but for NS, pruning the same number of channels will result in removing the entire layer for some individual experiment, so a relatively lower percentage of channels are pruned. From the result we don't see a distinct difference in the architecture stability between OT and NS. By pruning the same or more channels, OT out-performs NS on both computational complexity and performance. Particularly, on CIFAR-100 we can see that by using NS, layer 2$\sim$7 are almost not pruned while layer 12 is pruned to only 5.27 channels in average, indicating there are both under- and over-pruning in this case.

\begin{table*}[t]
	\label{table:statistics}
	\begin{center}
		\begin{tabular}{c|c|c|c|c|c} 
			\toprule[2pt]
			& OT,CIFAR-10 & NS,CIFAR-10 & OT,CIFAR-100 & NS,CIFAR-100 & Baseline \\
			\midrule[0.5pt]
			Layer 1 & 25.82$\pm$1.75 & 30.45$\pm$3.58 & 32.55$\pm$2.71 & 37.91$\pm$2.87 & 64 \\
			Layer 2 & 58.91$\pm$2.47 & 62.09$\pm$1.93 & 63.82$\pm$0.39 & 64.00$\pm$0.00 & 64 \\
			Layer 3 & 114.36$\pm$3.08 & 122.91$\pm$7.51 & 121.55$\pm$2.43 & 127.09$\pm$1.08 & 128 \\
			Layer 4 & 119.55$\pm$3.42 & 125.91$\pm$3.48 & 127.36$\pm$0.77 & 128.00$\pm$0.00 & 128 \\
			Layer 5 & 206.18$\pm$11.08 & 247.27$\pm$6.31 & 245.09$\pm$2.15 & 255.45$\pm$0.89 & 256 \\
			Layer 6 & 172.27$\pm$9.90 & 237.73$\pm$5.66 & 245.27$\pm$2.45 & 255.73$\pm$0.45 & 256 \\
			Layer 7 & 127.64$\pm$6.29 & 198.73$\pm$8.72 & 236.82$\pm$3.41 & 254.82$\pm$1.03 & 256 \\
			Layer 8 & 97.82$\pm$6.74 & 100.91$\pm$12.18 & 331.73$\pm$8.54 & 482.45$\pm$4.64 & 512 \\
			Layer 9 & 55.73$\pm$2.86 & 26.82$\pm$8.39 & 287.09$\pm$5.50 & 417.27$\pm$7.25 & 512 \\
			Layer 10 & 38.09$\pm$1.16 & 9.55$\pm$4.64 & 193.55$\pm$4.21 & 82.91$\pm$10.43 & 512 \\
			Layer 11 & 26.64$\pm$3.26 & 8.73$\pm$4.33 & 165.09$\pm$9.36 & 16.82$\pm$4.09 & 512 \\
			Layer 12 & 32.18$\pm$3.95 & 6.36$\pm$5.69 & 198.64$\pm$9.23 & 5.27$\pm$2.56 & 512 \\
			Layer 13 & 56.64$\pm$5.77 & 88.64$\pm$9.86 & 268.27$\pm$6.76 & 389.27$\pm$7.86 & 512 \\
			Total & 1131.82$\pm$46.76 & 1266 & 2516.82$\pm$29.94 & 2517 & 4224 \\
			FLOPs & \textbf{113.13M$\pm$6.84M} & 154.99M$\pm$8.06M & \textbf{198.10M$\pm$2.59M} & 226.68M$\pm$2.02M & - \\
			Params & \textbf{1.16M$\pm$0.09M} & 1.68M$\pm$0.04M & \textbf{4.74M$\pm$0.12M} & 5.02M$\pm$0.04M & - \\
			Accuracy & \textbf{93.52\%$\pm$0.12\%} & 93.39\%$\pm$0.17\% & \textbf{73.06\%$\pm$0.19\%} & 72.04\%$\pm$0.71\% & - \\
			\bottomrule[2pt] 
		\end{tabular}
	\end{center}
	\caption{Architectures, FLOPs, params, and accuracies of pruned VGG-14 using OT and NS. The results are obtained from 11 runs with random seeds. In each run we follow the pipeline: 1) Train with sparsity; 2) Prune; 3) Fine-tune. M is $10^6$.}
	\label{table:statistics}
\end{table*}

We also compared the method by Ye \emph{et al.}~\cite{ye2018rethinking} which is similar to NS and our method. We followed the experimental settings for ResNet-20 on CIFAR-10 from the paper. The number of training epochs is set to include the same iterations of both warm-up and training stages. For fine-tuning, we only use less than 1/10 of the iterations (12 epochs) compared to \cite{ye2018rethinking} because it is enough for the models pruned by OT to converge. We compare the absolute performance degradation with the number of parameters using the data reported in \cite{ye2018rethinking} in Figure \ref{fig:iclr2018}. Note that there is a slight difference between our reproduced baseline (91.85\%) and the baseline reported in the paper (92.0\%). From Figure \ref{fig:iclr2018}, regardless of whether the difference is considered, our method shows a better performance compared to \cite{ye2018rethinking}.

\subsection{Results on ILSVRC-2012}

\begin{table}[t]
	\begin{center}
		\setlength{\tabcolsep}{2.5mm}
		\begin{tabular}{c|c|c|c} 
			\toprule[2pt]
			Architecture & Method & Accuracy (\%) & GFLOPs\\
			\midrule[1pt] 
			\multirow{3}*{VGG-16-BN} & Baseline & 74.09 & 15.66\\
			& NS ($p$=0.415) & 65.47 / 68.66 / 71.02 & 8.33 \\
			& OT ($\lambda$=2e-4) & 72.02 / 71.98 / 72.16 & 7.71 \\
			
			\midrule[0.5pt]
			\multirow{3}*{PreResNet-50} & Baseline & 75.04 & 4.14 \\
			& NS ($p$=0.375) & 63.32 / 66.79 / 69.60 & 1.95 \\
			& OT ($\lambda$=2e-4) & 69.56 / 70.01 / 70.40 & 1.84 \\
			
			\bottomrule[2pt] 
		\end{tabular}
	\end{center}
	\caption{Results on ILSVRC-2012. $p$ represents the percentage of pruned channels for NS. In the column of ``Accuracy'', for NS and OT, from the left to right are the results fine-tuned by 1 epoch, 3 epochs and all the 15 epochs respectively. M and G represent $10^6$ and $10^9$ respectively}
	\label{table:ilsvrc2012}
\end{table}

\begin{figure}[t]
	\begin{center}
		\includegraphics[width=0.9\linewidth]{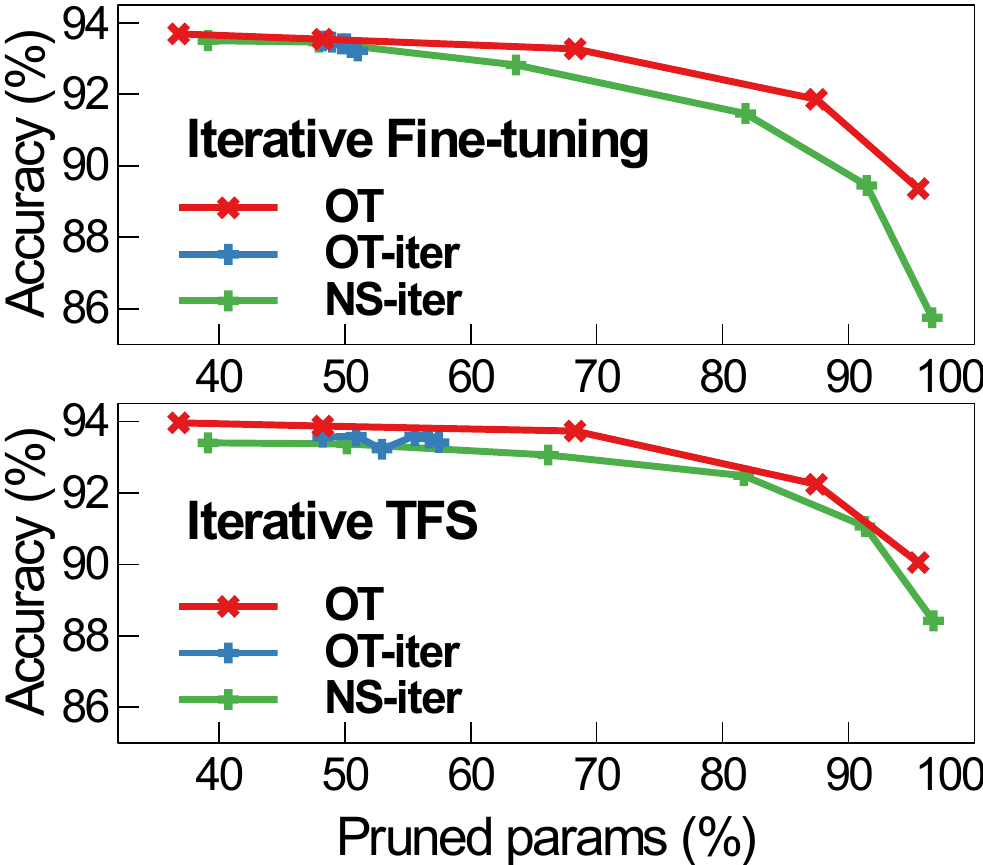}
	\end{center}
	\caption{Results of iterative pruning, and comparison with OT without iterative pruning. OT represents the result by one-shot pruning. OT-iter and NS-iter are results by iteratively pruning using OT and NS respectively.}
	\label{fig:iterative_prune}
\end{figure}

The results on ILSVRC-2012 is summarized in Table \ref{table:ilsvrc2012}. We can see that for both VGG-16-BN and PreResNet-50, OT outperforms NS at the same or even less computation budgets. Particularly, we find that by using OT, for VGG-16-BN the accuracy after only 1 epoch fine-tuning almost achieves the best accuracy, and for PreResNet-50 the accuracy after 3 epochs almost achieves the best accuracy. While by using NS the results after a few epochs fine-tuning are not good enough. For large scale data like ILSVRC-2012, OT significantly saves time for fine-tuning compared to NS. 

\subsection{Iterative Pruning}
Iterative pruning is a typical procedure in pruning~\cite{Liu_2017_ICCV,frankle2018the,liu2018rethinking}. It is very time consuming, but it often outperforms one-shot pruning~\cite{frankle2018the} or extend the percentage of pruned FLOPs to the region that cannot be achieved by one-shot pruning~\cite{Liu_2017_ICCV}. Interestingly, in this section we find that iterative pruning is not suitable for OT; however, we also show that OT outperforms iteratively pruned model by NS, even without the procedure. We conduct iterative pruning for VGG-14 on CIFAR-10. For NS we follow the original paper setup by pruning 50\% channels each time and pruning at most 50\% channels at each layer. For OT we keep $\lambda$=1e-4. 

In the upper graph of Figure \ref{fig:iterative_prune}, we can see that by iteratively using OT, the fine-tuned model almost cannot be further pruned from the second iteration. The probable reason may be that by using OT, the pruned model is very stable and does not lose important weights. We further explored the case of iteratively TFS using OT in the lower graph. Although it can prune a little more channels in each iteration than fine-tuning, the pruned percentage is still very small. This may imply that the architecture obtained by OT is very stable under a certain level of $L_1$ regularization. Although it is not a good idea to use OT iteratively, we find in Figure \ref{fig:iterative_prune} that even with only one-shot, OT outperforms the time-consuming iterative pruning using NS for both fine-tuning and TFS. 

\begin{figure}[t]
	\begin{center}
		\includegraphics[width=0.9\linewidth]{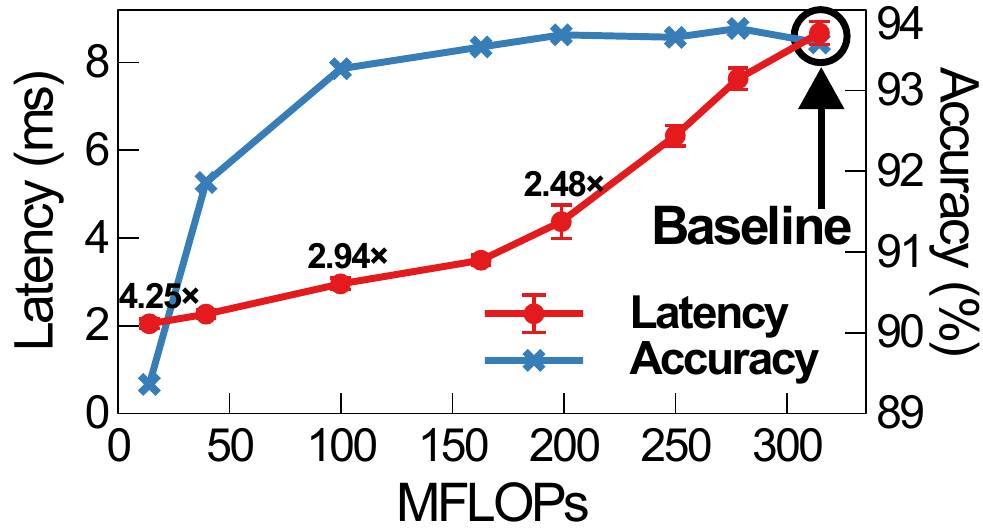}
	\end{center}
	\caption{Latencies and accuracies of pruned models on Intel NCS 2.}
	\label{fig:benchmark_ncs2}
\end{figure}

\subsection{Benchmark on Edge Accelerator}

We further benchmarked our method on a real edge computing accelerator since FLOPs does not always reflect the actual latency. We use Intel Neural Compute Stick (NCS) 2 for experiments. The models are VGG-14 trained on CIFAR-10. For each pruned model, we run single image inference for 10,000 iterations to obtain the average and standard deviation of latencies. The results with corresponding accuracies are shown in Figure \ref{fig:benchmark_ncs2}. The pruned model achieves on average 2.48$\times$ speedup without accuracy loss, and 4.25$\times$ speedup with $<5\%$ relative accuracy loss.

\section{Conclusion}

We have presented channel pruning design by layer-wise calculating the optimal thresholds to avoid both under- and over-pruning. To find the optimal threshold, we proposed an algorithm that preserves the cumulative sum of squares of the scaling factors from the BN layer. Our method is extremely simple yet effective and has been validated to outperforming state-of-the-art designs in extensive experiments. For the first time, the distribution of scaling factors in pruning and its relationship to under- and over-pruning is carefully studied by this work. We hope that further research based on our work can provide inspiration for finding more efficient networks.

\bibliographystyle{IEEEtran}
\bibliography{egbib}

\end{document}